# Retrieval Augmented Generation-based Large Language Models for Bridging Transportation Cybersecurity Legal Knowledge Gaps

**Khandakar Ashrafi Akbar\***
Research Assistant
Data Mining Lab
The University of Texas at Dallas,
800 W Campbell Rd, Richardson, TX 75080
Email: ashrafi@utdallas.edu

**Md Nahiyan Uddin**
Research Assistant
Data Mining Lab
The University of Texas at Dallas,
800 W Campbell Rd, Richardson, TX 75080
Email: md.uddin@utdallas.edu

**Latifur Khan, Ph.D.**
Professor, Department of Computer Science
The University of Texas at Dallas,
800 W Campbell Rd, Richardson, TX 75080
Email: lkhan@utdallas.edu

**Trayce Hockstad, J.D., M.A.**
Law & Policy Analyst, Transportation Policy Research Center
Alabama Transportation Institute, The University of Alabama
2031 SCIB, Tuscaloosa, Alabama, 35487
Email: tahockstad@ua.edu

**Mizanur Rahman, Ph.D.**
Assistant Professor in Transportation Systems Engineering
Department of Civil, Construction & Environmental Engineering
The University of Alabama
2007 SCIB, Box 870205, 248 Kirkbride Lane, Tuscaloosa, AL 35487
Email: mizan.rahman@ua.edu

**Mashrur Chowdhury, Ph.D.**
Professor, Automotive Engineering, Civil Engineering, and Computer Science
Clemson University
Glenn Department of Civil Engineering | Lowry Hall, Clemson, SC 29634
Email: mac@clemson.edu

**Bhavani Thuraisingham, Ph.D.**
Professor, Department of Computer Science
The University of Texas at Dallas,
800 W Campbell Rd, Richardson, TX 75080
Email: bxt043000@utdallas.edu

\*Corresponding author

# ABSTRACT

With the advancement of connected and automated transportation systems, a growing need has emerged for regulatory authorities to update existing laws and statutes, as well as create new ones, to address future implications of connectivity and automation. Specifically, ensuring proper engagement with cybersecurity and data privacy challenges for connected and automated transportation systems will require a comprehensive legal framework at both the federal and state levels. To help policymakers achieve this, a Retrieval Augmented Generation (RAG) based Large Language Model (LLM) framework has been developed in this study, focusing on extracting pertinent information from existing legislation based on inquiries and crafting LLM-generated responses to highlight potential loopholes for further scrutiny. This study primarily aims to mitigate the hallucinations caused in the domain of LLMs by generating a curated dataset of questions for enhanced training purposes. This RAG-based framework extracts relevant information to improve the specificity of answers and aids the LLM in providing factually accurate responses with increased reliability. Our analyses reveal that the presented RAG-based framework can aid legislative analysis by generating queries for particular questions and responses. We also compare our RAG framework-generated responses with commercially available LLMs to demonstrate the effectiveness of our approach. The RAG-powered LLM outperforms leading commercial LLMs across four distinct metrics, i.e., AlignScore, ParaScore, BERTScore, and ROUGE score. This highlights that integrating RAG allows LLMs to produce more factually accurate and up-to-date responses than standalone LLMs. This approach to domain-specific LLM development will improve the quality of legislative analysis that can be used to aid policymakers in meeting the challenge of updating legal codes in accordance with emerging technologies.

## 1. INTRODUCTION

The rise of connected and automated transportation systems underscores the pressing need to address legal uncertainties at the federal, state, and international levels. The intricate nature of these systems—particularly their reliance on networks of Internet of Things (IoT) devices—raises complex questions around data privacy, infrastructure integrity, and public safety [1]. Meaningful risk mitigation requires a careful examination and ongoing adaptation of legal frameworks across jurisdictions to account for emerging deficiencies. Large Language Models (LLMs) offer a critical tool in navigating this evolving intersection of technology and law. Their capacity for structured reasoning and evidence-based insight makes them uniquely suited to this task. With access to current legislative developments, these models enable comprehensive analyses aimed at identifying legal gaps and ensuring that regulatory responses remain aligned with the pace of technological change in the transportation sector [2].

The legislative landscape, defined by its precision and specificity, demands close attention and leaves little margin for error. As legislative bodies undertake amendments and introduce new policy frameworks, their understanding of the current legal environment must be both accurate and detailed. Misrepresentation or omission of critical information can lead to significant delays and financial costs associated with restructuring legal or regulatory systems [3]. With these concerns in mind, we examine the limitations of current commercial LLMs in accurately capturing and interpreting prevailing standards of governance in the United States (U.S.). Leveraging LLMs to understand better the law can benefit legal stakeholders [4], expedite legal research [5], and help technology stakeholders align their policies or regulations with federal, state, and international laws.

Incorporating new legislation into LLMs is vital for accurate legal analysis, such as transportation cybersecurity, which could profoundly impact public safety and security. However, retraining these models with each legislative update is impractical due to the length and complexity of legal codes [6]. When pre-trained LLMs generate responses based on recent developments, such as newly passed laws, they often produce hallucinations, as shown in Figure 1. This issue (explained in an exemplary way in Figure 1) mainly occurs because the queries do not match the knowledge available in the LLMs during their training time. Retrieval Augmented Generation (RAG) offers a promising solution within Natural Language Processing (NLP) by efficiently retrieving relevant information based on the given context or prompts.

RAG is a combination of retrieval and generation-based processes. It extracts pertinent information from the external database to enhance the quality of the generated responses considering the contexts. In RAG, the retriever fetches relevant documents from a curated knowledge base, and the generator leverages them to create context-aware responses, which enhances answer quality by grounding outputs in accurate, current information, minimizing hallucinations. Fine-tuning large language models (LLMs) for downstream tasks often presents significant challenges and may not consistently deliver optimal outcomes [6]. Given these limitations and the increasing importance of addressing legal concerns in the transportation sector, our approach offers a compelling alternative. It allows for the effective handling of targeted queries that reveal legislative knowledge gaps—without requiring extensive retraining of the underlying models. The objective of this study is to leverage the RAG framework to answer queries that can reveal these legislative knowledge gaps accurately. This method eliminates the need for retraining LLMs with insufficient data and improves the factual accuracy and consistency of the generated responses.

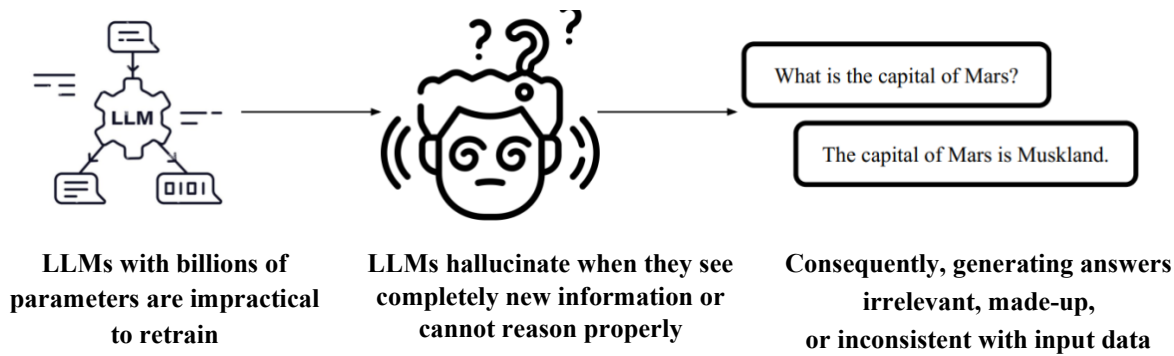

Figure 1: Hallucination in Large Language Models

## 2. CONTRIBUTION OF THE RESEARCH

The primary contribution of this research is tackling the hallucination issue prevalent in commercial state-of-the-art LLMs when generating responses in specialized and rapidly evolving domains, such as transportation cybersecurity. LLM hallucination in legal contexts is critical due to: 1) the sensitivity of legal matters demanding heightened accuracy, and 2) the rapidly evolving legal landscape requiring adaptable technologies. Specifically, RAG mitigates LLM hallucinations by supplying the latest context for informed responses. While commercially deployed LLMs may struggle with quick retraining on newly enacted laws, RAG significantly accelerates this adaptation, ensuring up-to-date and reliable outputs.

## 3. RELATED WORK

To contextualize our study, this literature review critically examines the development of LLMs with a focus on hallucination—a significant limitation that undermines their reliability. We explore how generative models have evolved, the nature and taxonomy of hallucinations, and the approaches taken to detect and mitigate them. Our goal is to identify unresolved challenges that motivate the contributions of this work, particularly within domain-specific and high-stakes applications like legal and policy reasoning.

The introduction of BERT [7] revolutionized natural language understanding through its bidirectional training and contextual embeddings, establishing a foundation for numerous language tasks. Specialized variants like ConfliBERT [8] adapted this architecture for nuanced areas, such as political conflict analysis, demonstrating the adaptability of pretrained models. However, with the rise of generative language models, a key challenge has emerged: hallucination—the generation of content that is incorrect, unfaithful to input, or unsupported by factual knowledge. Sun et al. [9] initiated important discussions around unsafe outputs in conversational artificial intelligence (AI), especially in sensitive contexts such as healthcare, and proposed DiaSafety, a benchmark for detecting such outputs. While useful, the work primarily focuses on conversational safety and not the root cause of factual inconsistency in complex domains.

Zhang et al. [10] and Huang et al. [11] offered taxonomies that categorize hallucinations into types, such as input-conflicting, context-conflicting, and fact-conflicting. These taxonomies clarify where hallucinations occur but stop short of offering concrete mitigation strategies, especially in domain-specific settings where factual correctness is crucial. Dziri et al. [12] traced hallucinations to poor dataset curation and model design, suggesting that architecture alone cannot guarantee truthfulness. Similarly, Huo et al. [13] explored self-verification using retrieval but found it insufficiently reliable. This reveals a gap in combining robust retrieval with domain grounding. Liu et al. [14] introduced HaDeS, targeting hallucination at the token level without needing full ground truth. While this granularity is valuable, it lacks applicability in domains like law or medicine, where reference precision and context are essential. In another study, Gao et al. [15] proposed ALCE, which evaluates LLMs' citation accuracy, highlighting that even when retrieval is used, models often generate content without verifiable sources. These insights underscore that retrieval alone is not enough—reasoning and synthesis from retrieved documents remain weak spots. This concern is echoed in domain-specific evaluations. A Stanford study [16] shows that models like GPT-3.5 and LLaMA 2 often hallucinate in legal tasks, misrepresenting legal facts and logic. Despite their general capabilities, these models falter in structured reasoning and reliable citation—both critical in legal or policy applications. Recent advances like Retrieval-Augmented Generation (RAG) [17] and RAFT [18] attempt to close this gap by combining generation with retrieval. RAG uses Wikipedia-scale data to ground responses, while RAFT adds chain-of-thought reasoning to enhance citation fidelity in biomedical and question answering domains. Despite outperforming traditional models, both approaches are general-purpose and not explicitly tailored to high-risk, specialized domains like law or cybersecurity policy.

Motivated by the issue of hallucination in LLMs, our work focuses on creating tailored datasets and RAG-enhanced LLMs to effectively identify legal loopholes in transportation policy. Accurate interpretation of the continuously evolving legal landscape is vital for detecting regulatory gaps and designing informed solutions. To this end, we develop a RAG-based LLM capable of seamlessly integrating ongoing legal updates across all levels, enabling consistent and reliable analysis of shortcomings within existing legal frameworks.

## 4. RETRIEVAL AUGMENTED GENERATION (RAG)

Sustaining LLMs presents significant challenges, primarily due to the enormous data requirements for training and fine-tuning and the computational resources needed. To mitigate these complexities, context-augmented LLMs offer a practical solution. Data frameworks such as LlamaIndex [19] facilitate the development of applications utilizing this approach. RAG is a prominent context-augmented language generation method that enhances the capability of LLMs to provide contextually relevant answers, even when such information is not explicitly present in the training dataset.

Figure 2 illustrates how RAG enhances LLMs by combining their static training knowledge with up-to-date external information. New information is stored in a vector database, retrieved based on user queries, and then used by the LLM to generate accurate, current responses. Responses generated by RAG-powered LLMs tend to be more specific, factually correct and varied in content. RAG frameworks can be optimized by updating the retriever and the generator or by keeping the retriever fixed and only updating the generator. When adapting to a new domain, especially from a linguistic perspective, jointly updating both components is often beneficial. In the foundational work on RAG [20], the retriever and the generator are trained end-to-end, with the retrieved documents treated as latent variables. The RAG-sequence and RAG-token models are developed by marginalizing over the latent documents, producing a distribution over the generated text. The key difference between these two models lies in handling retrieved documents during generation. In the RAG-sequence model, distinct output sequences are generated for each retrieved document, which are subsequently marginalized. In contrast, the RAG-token model allows the generator to select content from different documents to generate the final answer. The marginalization process in both approaches enhances the robustness of the RAG framework to retrieval noise, as some retrieved documents may be only partially relevant or irrelevant to the query. Despite technical challenges in implementing RAG—such as limitations on the number of retrieved documents and the chunk length—theoretical advancements enable improvements that mitigate these issues, delivering reasonable results without heavily relying on computational resources.

In the domain of legal analysis, where legislative integration is a continuous process across various bodies (federal, state, and international), RAG presents a particularly effective solution. This approach can manage extensive legal knowledge while ensuring high contextual relevance. Furthermore, RAG enables rapid adaptation to new legislative bodies or updated versions of existing legislation. Therefore, the use of context-augmented generation processes like RAG is instrumental in navigating the evolving legal landscape, ensuring that analyses remain current and comprehensive.

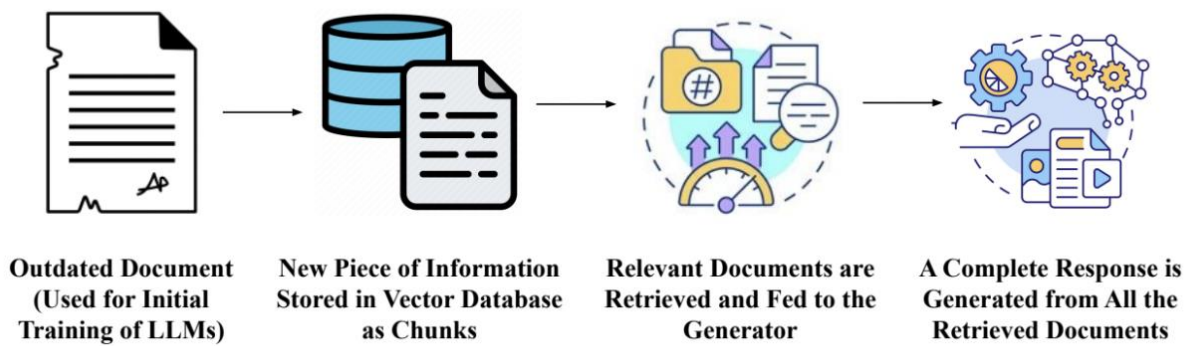

**Figure 2: Architectural Overview of RAG**

## 5. SYSTEM OVERVIEW

In this section, we present the RAG pipeline. This pipeline employs two distinct methods for relative context retrieval. The first method generates the retrieval index on the entire dataset. When a question is posed to the LLM, the retriever searches through all indices to find the most relevant ones, subsequently generating the answer. Although this process is more time-consuming, it is generally effective when the question does not specify any state's name. Figure 3 illustrates the natural process for the first method. Initially, a query is submitted to the context-augmenting response generator (RAG-powered LLM). The retriever generates indices for all the states and searches through them to find the most relevant information corresponding to the query. Finally, the output is produced. Figure 4 illustrates the second method, where a dedicated retriever is assigned to each individual state. The specific states mentioned are retrieved first when a query involves multiple states. Each retriever then searches through the relevant state indices and generates a response for each state. The final response is produced by combining these individual responses.

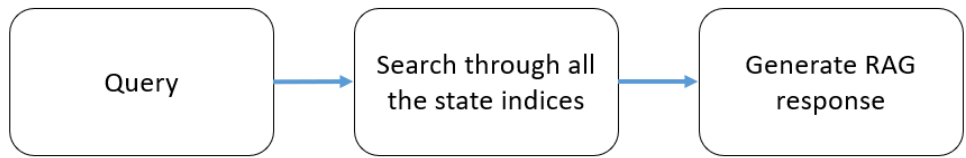

Figure 3: Flow of Method 1 (Index on Entire Data)

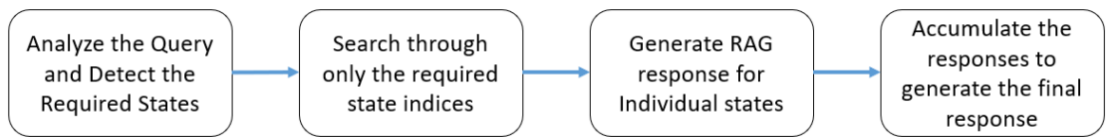

Figure 4: Flow of Method 2 (State Wise Individual Retriever)

## 5.1 WHOLE DATA INDEX (WDI)

The first method in developing the retrieval augmented generation pipeline works by searching the entire index database. We name this method 'Whole Data Index (WDI)' as it searches through the indexes of all the states' data. As depicted in Figure 5, the laws we source from different legislatures are used to populate or generate a vector database with the indexes of the pieces of legislation. Different legislations from the state level are first collected as documents and split into chunks (e.g., a single document will contain multiple chunks). These chunks serve as the node in the vector database and embeddings for each node are generated and stored. When a query is presented to the RAG framework, this entire database is searched through to find the top k (k is variable here, e.g., it can take a value of 5, 10, 15, and so on) relevant chunks of documents to answer that query. Then, those chunks, along with the question, are given to LLM to generate the final response to that query.

## 5.2 STATE-WISE INDEX (SWI)

The second method we employ for generating responses using the LLMs employs only the relevant state chunks or indexes. We call this approach 'State-Wise Index (SWI)' as it only looks through the indexes of the states mentioned in the query. This development of the retrieval augmented generation pipeline is a clever solution to the searching problem because when the queries involve the name of the states, the search can only be kept limited to a certain number of states. First, a query that mentions a single state or multiple states is analyzed to extract relevant state documents to look at. Only the chunks relevant to those states are then considered while extracting the relevant chunks of that query. For example, let's say a query mentions the name of the state of "Alabama" and the state of "Kansas" in the query. First, the two mentioned state names will be extracted from the query. Afterward, the documents and further chunks from these two states will be searched for the most relevant nodes or indexes given that query. The extracted chunks from those documents serve as the components for the vector database. These chunks are ultimately represented as nodes, a structure designed to enhance the efficiency of searches by organizing and structuring the data within the vector database. Then, each state's top k relevant chunks will be retrieved as shown in Figure 6. These top k elements from each state and a prompt will be provided to the LLM, and a response will be generated. This type of response will be generated for all participating states in the query. Then, all the state-wise responses will be accumulated, and the final response will be generated for that query. This method is particularly effective for generating responses to queries that involve multiple states. The advantage of this approach is further elaborated in section 7.2.

## 5.2 TECHNOLOGY SPECIFICATION

We have developed a web-based interface where specific questions can be asked to our retriever-generator model. The basic technology stack used for these two methods is provided in Table 1. The embedding model encodes text information from legislative documents into embeddings, which are stored in a vector database. Indices are created to facilitate optimal searching through these embeddings. When a question is asked, it is passed to the retriever, which establishes an embedding of the question and retrieves similar text chunks using similarity search. These texts are provided to a generator model, which uses them as context along with the question to generate a response. This entire process is hosted on a web interface, where users can input questions. Based on the question, the response is generated on the backend, and the processed responses are displayed on the frontend.

**Table 1: Technology Specification**

| Component | Version/Framework |
|---|---|
| Retriever Embedding Model | text-embedding-ada-002 (OpenAI) |
| Generator Model | gpt-3.5-turbo (OpenAI) |
| Additional Library | LlamaIndex |
| Web-Interface Backend | Django |
| Web-Interface Front-End | HTML, CSS, Javascript, Bootstrap |

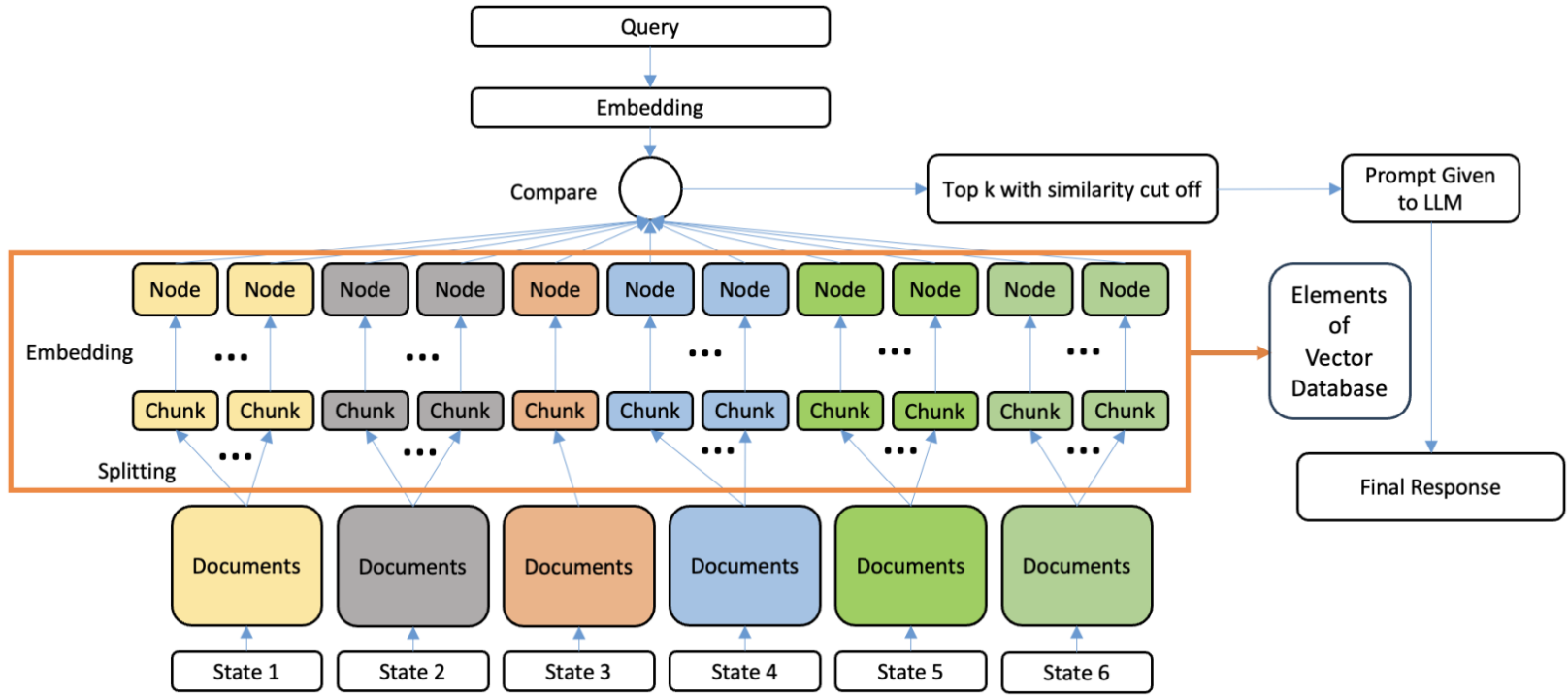

**Figure 5: Complete Architecture for Index on Entire Data: Whole Data Index (WDI)**

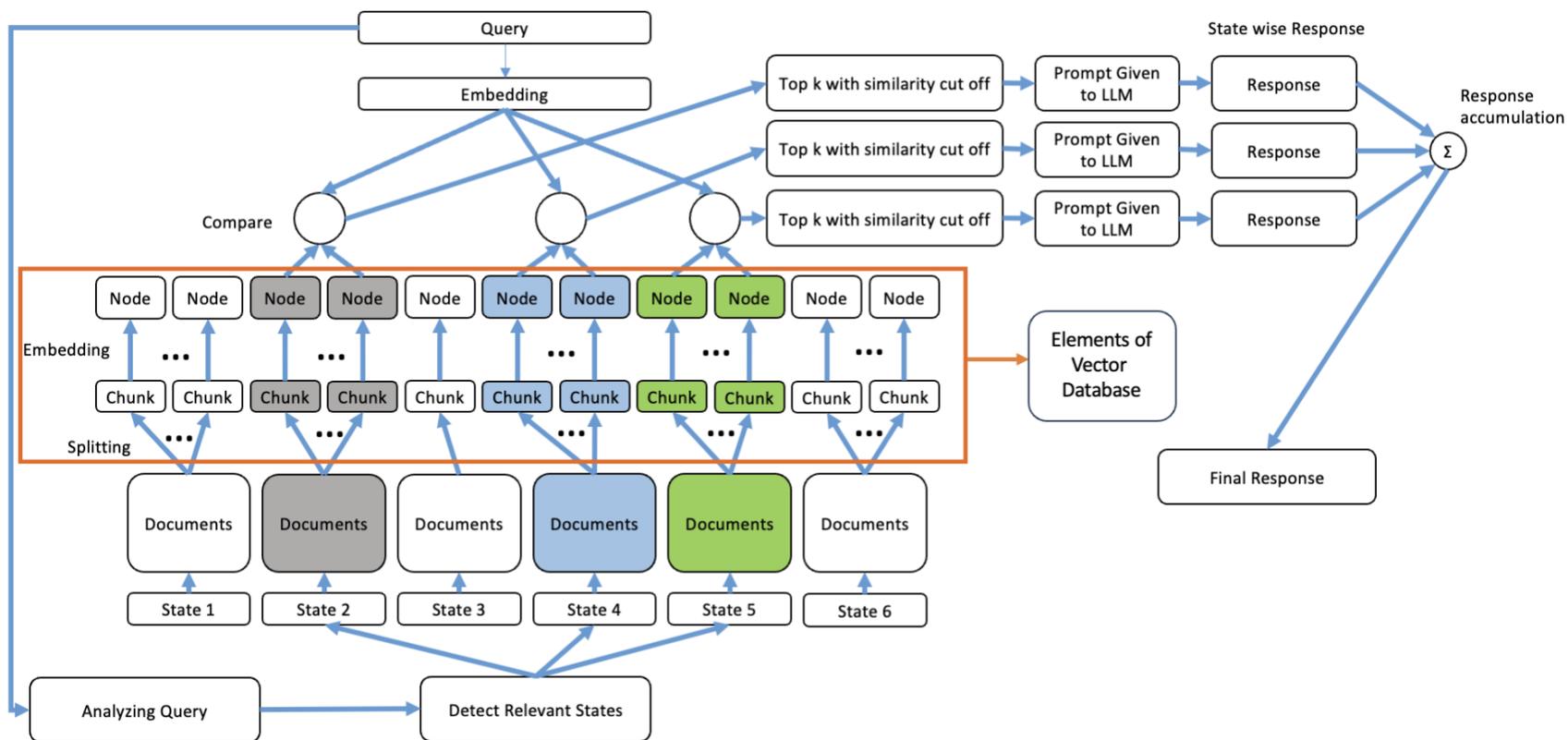

**Figure 6: Complete Architecture for State-Wise Individual Retriever: State-Wise Index (SWI)**

## 6. DATASET PREPARATION

To develop a comprehensive dataset of currently enacted U.S. state-level legislation relevant to cybersecurity governance, a systematic review of all 50 state statutory codes was conducted. The purpose of this exercise was to identify, categorize, and extract legislative provisions that address core areas of cybersecurity law and policy across state jurisdictions. All searches were performed by a licensed attorney using the LexisNexis legal research platform, ensuring that the results reflect authoritative legal texts currently in force.

For each state, the search was conducted individually within the "statutes and legislation" content category. The first search term used was "cybersecurity", entered without quotation marks. Results were limited to the official "Code of [State]" for each jurisdiction, with all available publication dates included. Relevant code sections containing substantive references to cybersecurity were then reviewed and selected for inclusion in the dataset.

A second targeted search was conducted using the term "data breach", again entered without quotation marks. The goal was to identify legislative provisions addressing breach notification obligations, consumer protection requirements, and related regulatory responses to data incidents. As with the previous search, results were limited to each state's code, all publication dates were included, and only unique, non-duplicative sections were retained.

The third search term used was "crim! use of computer", leveraging LexisNexis's wildcard operator to capture variations, such as "criminal use of computer," "criminalized use of computer," or similar statutory constructions. This search focused on identifying criminal statutes implicating unauthorized access, malicious activity, or fraudulent use of computer systems. Filtering and selection parameters remained consistent with those described above.

Subsequent searches were conducted for "denial of service" and "ransomware," respectively, to isolate legislative language addressing network disruption and extortion-based cybercrime. As these are emerging areas of state-level criminalization, this portion of the review emphasized the identification of specialized offenses as distinct from general computer crime statutes.

Finally, each state's code was searched for "trade secrets," which served as a proxy for identifying cybersecurity-relevant statutes that protect proprietary digital information or regulate access to sensitive technological assets. Where applicable, citations to each state's Uniform Trade Secrets Act (UTSA) or analogous statutory framework were included.

Across all search terms, the following method was consistently applied: results were filtered by jurisdiction, limited to the "Code of [State]" source category, and inclusive of all publication dates available in LexisNexis. Relevant code sections were selected on the basis of substantive content and non-duplication within the dataset. Certain search results, while topically adjacent to cybersecurity, were excluded from the dataset on the grounds that they fell outside the scope of this research. These included statutes related to election security, cyberbullying, cyberstalking, taxation, and child sexual exploitation (CSE/C) laws. These areas, although important, were

## 6.1 COMPILATION OF LEGISLATIVE DOCUMENTS

Having meticulously compiled federal and state-level legislations concerning cybersecurity and data privacy in connected and automated transportation systems, we employ these statutes as foundational pillars for constructing our RAG pipeline. They serve as essential contextual cues for generating responses from our curated dataset. State statutes are meticulously gathered, organized by the state, and categorized by topic. Following the establishment of this fundamental framework for our legal landscape analysis, we shift our focus to creating a dataset of customized questions to be used with our RAG pipeline. In designing this dataset, we prioritize both legal style and technological perspectives. Below are a few exemplary questions drawn from the dataset: (i) If there is any trace of a cyber-attack by hijacking sensor inputs/values, who might be held liable in the event of a breach? and (ii) Do the regulations provide sufficient guidance on encryption standards, data encryption requirements, and secure data storage practices? In Figure 7, we illustrate the steps in our workflow, beginning with the collection of legislation from various sources. This collected data is then used to populate a vector database, which serves as the foundation for our RAG pipeline to generate responses. These generated responses play a crucial role in identifying legislative gaps.

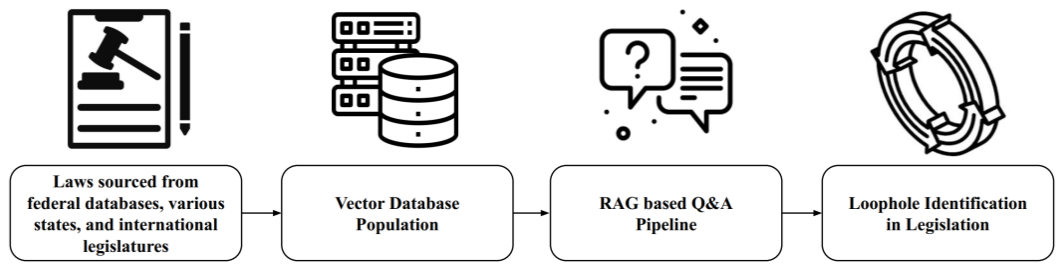

**Figure 7: Workflow Diagram**

## 6.2 SAMPLE DATA SNIPPET

Figure 8 demonstrates a sample of the Code of Alabama. The data snippet describes the § 13A-8-112 which addresses the crime of computer tampering, defining it as unauthorized access to, alteration of, or destruction of computers, computer systems, or networks. This document, along with many others like it from other jurisdictions was indexed in the dataset such that these legal terms and definitions are now information that can be queried through the RAG system. As such, the LLM can provide information about the legal terminology contained in the statute.

> # Code of Ala. § 13A-8-112
>
> Current through the end of the 2023 First Special, Regular, and Second Special Sessions, but not including corrections and changes made to the 2023 session laws by the Code Commissioner.
>
> *Michie's™ Alabama Code > TITLE 13A Criminal Code (Chs. 1 — 14) > CHAPTER 8 Offenses Involving Theft (Arts. 1 — 11) > Article 5A The Alabama Digital Crime Act (§§ 13A-8-110 — 13A-8-119)*
>
> ## § 13A-8-112. Computer tampering.
>
> (a) A person who acts without authority or who exceeds authorization of use commits the crime of computer tampering by knowingly:
>
> (1) Accessing and altering, damaging, or destroying any computer, computer system, or computer network.
>
> (2) Altering, damaging, deleting, or destroying computer programs or data.
>
> (3) Disclosing, using, controlling, or taking computer programs, data, or supporting documentation residing in, or existing internal or external to, a computer, computer system, or network.
>
> (4) Directly or indirectly introducing a computer contaminator or a virus into any computer, computer system, or network.
>
> (5) Disrupting or causing the disruption of a computer, computer system, or network services or denying or causing the denial of computer or network services to any authorized user of a computer, computer system, or network.
>
> (6) Preventing a computer user from exiting a site, computer system, or network-connected location in order to compel the user's computer to continue communicating with, connecting to, or displaying the content of the service, site, or system.
>
> (7) Obtaining any information that is required by law to be kept confidential or any records that are not public records by accessing any computer, computer system, or network that is operated by this state, a political subdivision of this state, or a medical institution.
>
> (8) Giving a password, identifying code, personal identification number, debit card number, bank account number, or other confidential information about a computer security system to another person without the consent of the person using the computer security system to restrict access to a computer, computer network, computer system, or data.
>
> (b)
>
> (1) Except as otherwise provided in this subsection, the offense of computer tampering is a Class A misdemeanor, punishable as provided by law. Subsection (a) does not apply to any acts which are committed by a person within the scope of his or her lawful employment. For purposes of this section, a person acts within the scope of his of her employment when he or she performs acts which are reasonably necessary to the performance of his or her work assignment.
>
> Trayce Hockstad

**Figure 8: Sample Data Snippet of the Legislation Documents (Code of Ala. § 13A-8-112)**

## 6.3 DATASET STATISTICS

Table 2 provides an overview of the dataset statistics compiled for our framework generation. It highlights the range of collected documents, the various sources of these legislations, and the corresponding document counts. This table breaks down the number of statutes and other legislative documents that have been compiled and indexed in the LLM's dataset. As shown, the dataset is expansive in coverage of legislative scope, with federal, international, and all fifty states represented in the archive. This compilation was put together selectively by a licensed attorney rather than automatically curated through internet searching.

**Table 2: Dataset Statistics**

| Source of Legislations | Number of Legislative Documents | Year Span |
|---|---|---|
| Federal | 86 | 2012-current |
| State | 1704 | All current enacted |
| International | 5 | 2018-current |
| Total | 1795 | 2012-current |

## 6.4 QUESTION AND ANSWER DATASET

The team of both legal and engineering professionals worked together over several weeks to select the 59 initial questions that were used to train the LLM. The questions were targeted to what appeared as most-discussed issues in transportation cybersecurity. The questions were also developed in relation to the legislation that exists in the U.S. A licensed attorney working on the project initiated the first draft of questions based on real-world legal queries covering cybersecurity issues. The attorney also consulted with other cybersecurity legal experts in the process of drafting the initial question list. Legal terminology and relevant definitions were considered in wording prompts so that the LLM would be able to interact most naturally with the statutory vocabulary. The questions were revised several times to ensure they were not duplicative and provided the best range of possible responses to gauge the LLM's performance. To demonstrate the effectiveness of our context-augmented language generation framework in the legal landscape, we have curated a question-and-answer dataset that redirects to scratching the legal surface for deeper understanding in the domain of transportation cybersecurity. We demonstrate some of the sample questions in Table 3. We have also curated the reference answers to these demonstrative questions that eventually serve as ground truths.

**Table 3: Sample Questions from the Q&A Dataset**

| Sample No | Question |
|---|---|
| #1 | Who might be liable if a passenger's credit card details were accessed without authorization while using services from an autonomous vehicle? |
| #2 | How do states address cybersecurity risks associated with cloud computing? |
| #3 | Are there any significant differences in the penalties or fines imposed for non-compliance with cybersecurity regulations among states? |
| #4 | What are the challenges faced by states in implementing and enforcing cybersecurity regulations, and how do they vary across jurisdictions? |
| #5 | Has California passed any legislation that references the Cybersecurity and Infrastructure Security Agency's Final Rule? |

## 7. RAG EVALUATION

We made great strides in resolving the problem of insufficiently trained LLMs lacking up-to-date information and the issue of hallucination by utilizing RAG. Figure 9 showcases a sample output produced by our RAG-powered GPT architecture. This architecture leverages RAG to transform datasets collected at federal, state, and international levels into indexed embeddings. When queried about specific contexts or laws, these embeddings are utilized to retrieve the most pertinent information along with their corresponding contexts and sources.

In Figure 9, the sources displayed are references from documents identified as highly relevant to the question, retrieved by the RAG architecture's retriever. These contextual snippets are combined with the user's query and an additional prompt to create a comprehensive input for the large language model (LLM). This process enables the LLM to generate coherent and accurate responses, effectively mitigating the risk of hallucinations.

Furthermore, the RAG architecture returns these sources alongside the generated responses (as shown at the bottom of Figure 9). This feature provides references, allowing users to verify the accuracy of the response or pinpoint the exact documents from which the information was sourced. This approach ensures transparency and precision in the generated outputs, making it easier to trust and validate the responses provided by the LLM.

---

**Sample Output:**

**You:**

Have any states pass legislation concerning cybersecurity in the context of autonomous vehicles?

**TraCR AI - Response:**

Yes, several states have passed legislation concerning cybersecurity in the context of autonomous vehicles. For example, in the context provided from Colorado Revised Statutes Annotated, there is a detailed section (43-1-130) that mandates a study on issues related to the development and adoption of autonomous motor vehicles. This study includes assessing safety benefits, risks (including cybersecurity risks), legal issues, as well as modifications needed for existing state transportation infrastructure to enable the use of autonomous vehicles. Additionally, in the context provided from Maryland Insurance Code Annotated, there are actions required following a cybersecurity event involving carriers, highlighting the importance of notifying the Commissioner promptly in case of cybersecurity incidents.

**Sources:**

C:\Users\User\Box\UTD\PhDResearch\RAG-app\CurrentCybersecurity Law\Colorado\Transportation\C.R.S. 43-1-130.pdf

C:\Users\User\Box\UTD\PhDResearch\RAG-app\CurrentCybersecurity Law\Maryland\Insurance\Md.

---

**Figure 9: Sample Result Generated by our RAG-powered GPT Architecture**

The architecture is designed to extract relevant information from a large set of source documents and conduct targeted comparisons. When evaluated against a set of reference answers, the system consistently produced responses that were semantically aligned. Table 4 below presents a selection of sample question-and-answer outputs. As shown, the RAG pipeline reliably generates responses that closely mirror the meaning and structure of the established ground truths. We have shown through our experimentation that our RAG-powered LLM performs better in answering complex, intricate queries. This indicates that our developed architecture can effectively handle a broader range of complex queries, and the generated responses can be used to visualize complex legal analysis. The synthesis of the framework to uncover legislative knowledge gaps can be effectively driven by a well-crafted set of questions. We focused on developing a solution that could address any question related to this aspect, ensuring alignment with base or reference answers. The next phase involves curating a more comprehensive set of questions to identify and highlight the existing loopholes in the legal landscape of autonomous and connected vehicles.

Table 5 presents sample responses generated by baseline LLMs (such as ChatGPT-4o, Claude, and Gemini 2.0 Flash). The purpose of this table is to illustrate the effectiveness of our proposed method through comparison. As of 2025, leading AI models like GPT-4o [21], Claude 3 Opus [22], and Gemini 1.5 Pro significantly outperform GPT-3.5-turbo in reasoning, code generation, and multimodal capabilities. These newer models offer vastly improved context windows (up to 1M tokens), faster performance, and greater accuracy across tasks.

Table 4: Case Study on the Effectiveness of RAG-powered LLM

| Question | Response from TraCR RAG | Reference Answer |
|---|---|---|
| What is the definition of identification document according to Code of Ala. § 13A-8-111? | Looking into the following state(s):<br>1. Alabama<br><br>Alabama:<br>According to Code of Ala. § 13A-8-111:<br>14) Identification document. — Any document containing data that is issued to an individual and which that individual, and only that individual, uses alone or in conjunction with any other information for the primary purpose of establishing his or her identity or accessing his or her financial information or benefits. Identification documents specifically include, but are not limited to:<br>a. Government issued driver's licenses or identification cards.<br>b. Payment cards such as credit cards, debit cards, and ATM cards.<br>c. Passports.<br>d. Health insurance or benefit cards.<br>e. Identification cards issued by educational institutions.<br>f. Identification cards for employees or contractors.<br>g. Benefit cards issued in conjunction with any government supported aid program.<br>h. Library cards issued by any public library.<br><br>References:<br>1) Current Cybersecurity Law\Alabama\Digital Crime Act\\_13A-8-111. Definitions.pdf<br>2) Current Cybersecurity Law\Alabama\Data Breach Notification Act\Files(12).pdf<br>3) Current Cybersecurity Law\Alabama\Digital Crime Act\\_13A-8-111. Definitions.pdf | **(14) Identification document.** — Any document containing data that is issued to an individual and which that individual, and only that individual, uses alone or in conjunction with any other information for the primary purpose of establishing his or her identity or accessing his or her financial information or benefits. Identification documents specifically include, but are not limited to, the following:<br>**a.** Government issued driver's licenses or identification cards.<br>**b.** Payment cards such as credit cards, debit cards, and ATM cards.<br>**c.** Passports.<br>**d.** Health insurance or benefit cards.<br>**e.** Identification cards issued by educational institutions.<br>**f.** Identification cards for employees or contractors.<br>**g.** Benefit cards issued in conjunction with any government supported aid program.<br>**h.** Library cards issued by any public library. |
| How do most states define critical infrastructure in the context of cybersecurity? | Most states (among all 50 states) define critical infrastructure in the context of cybersecurity as systems and assets, whether physical or virtual, so vital to the state that their incapacity or destruction would have a debilitating impact on security, economic security, public health, safety, or any combination thereof. | Critical infrastructure means existing and proposed information technology and operational technology systems and assets, whether physical or virtual, the incapacity or destruction of which would negatively affect security, economic security, public health, or public safety. |
| Has California passed any legislation that references the Cybersecurity and Infrastructure Security Agency's Final Rule? | Looking into the following state(s):<br>1. California<br><br>California:<br>I am sorry, I could not find any information to answer the question you asked. | None. |

Table 5: Case Study on the Effectiveness of Baseline LLMs

| Question | Response from TraCR RAG | Reference Answers |
|---|---|---|
| What is the definition of identification document according to Code of Ala. § 13A-8-111? | Looking into the following state(s): 1. Alabama<br><br>Alabama: According to Code of Ala. § 13A-8-111:<br>14) Identification document. — Any document containing data that is issued to an individual and which that individual, and only that individual, uses alone or in conjunction with any other information for the primary purpose of establishing his or her identity or accessing his or her financial information or benefits. Identification documents specifically include, but are not limited to:<br>a. Government issued driver's licenses or identification cards.<br>b. Payment cards such as credit cards, debit cards, and ATM cards.<br>c. Passports.<br>d. Health insurance or benefit cards.<br>e. Identification cards issued by educational institutions.<br>f. Identification cards for employees or contractors.<br>g. Benefit cards issued in conjunction with any government supported aid program.<br>h. Library cards issued by any public library.<br><br>References:<br>1) Current Cybersecurity Law\Alabama\Digital Crime Act\_13A-8-111. Definitions.pdf<br>2) Current Cybersecurity Law\Alabama\Data Breach Notification Act\Files(12).pdf<br>3) Current Cybersecurity Law\Alabama\Digital Crime Act\_13A-8-111. Definitions.pdf | **ChatGPT-4o:**<br>Sorry, I can't guarantee accurate information about voter-related laws and procedures in the U.S., including Code of Ala. § 13A-8-111. I recommend checking the official Alabama state legislature website or consulting a legal expert for the most accurate information.<br><br>**Claude:**<br>For the specific legal definition of an identification document according to the Code of Alabama § 13A-8-111, I would need to reference the exact text of the statute.<br><br>Based on Alabama Code § 13A-8-111, an "identification document" typically refers to a document made or issued by or under the authority of the United States government, the government of another state, political subdivision of another state, a foreign government, political subdivision of a foreign government, an international governmental or quasi-governmental organization which, when completed with information concerning a particular individual, is of a type intended or commonly accepted for the purpose of identification of individuals.<br><br>However, to give you the precise legal definition as currently written in Alabama law, I recommend consulting the official Alabama Code through the Alabama Legislature's website or speaking with a legal professional who can provide the exact statutory language. The specific definition may include additional details or exceptions that would be important for legal purposes.<br><br>**Gemini (Version: 2.0 Flash):**<br>According to Code of Ala. § 13A-8-111(14), an "identification document" is defined as:<br><br>Any document containing data that is issued to an individual and which that individual, and only that individual, uses alone or in conjunction with any other information for the primary purpose of establishing his or her identity or accessing his or her financial information or benefits.<br><br>The statute further specifies that identification documents specifically include, but are not limited to, the following:<br><br>* Government issued driver's licenses or identification cards.<br>* Payment cards such as credit cards, debit cards, and ATM cards.<br>* Passports.<br>* Health insurance or benefit cards.<br>* Identification cards issued by educational institutions.<br>* Identification cards for employees or contractors.<br>* Benefit cards issued in conjunction with any government supported aid program.<br>* Library cards issued by any public library. |

# 8. COMPARATIVE EVALUATION OF RAG-POWERED GPT ARCHITECTURE

We compare the performance of our RAG-powered LLM with other state-of-the-art commercial LLMs in generating responses for a subset of our Q&A dataset. The comparison focuses on the factual accuracy of the responses produced by each model. The more accurate and factually consistent a response is with the ground truth, the more successful the model is considered.

## 8.1 BASELINE APPROACHES

This section offers a glimpse into several of the leading commercial large language models (LLMs).

### 8.1.1 CHATGPT
ChatGPT [23], launched by OpenAI in November 2022, is a chatbot powered by large language models (LLMs) that allow users to customize conversations. Its rapid success, with over 100 million users by January 2023, contributed to an AI boom, leading to the creation of competing products and the integration of ChatGPT into Apple's services. While praised for its capabilities, it has also sparked concerns about its impact on human intelligence and potential misuse. ChatGPT, based on OpenAI's GPT models, began as a free research preview and now operates on a freemium model, offering additional features through subscriptions. The versions of ChatGPT used for this baseline study were both ChatGPT3.5 and ChatGPT4o.

### 8.1.2 CLAUDE
Claude [24], developed by Anthropic, is a family of large language models first released in March 2023. These models, including the latest Claude 3.5 Sonnet, use Constitutional AI, a training method designed to make them helpful, honest, and harmless. Claude's iterations have introduced features like image analysis, expanded token capacity, and advanced reasoning capabilities. The models undergo a unique training process where they self-critique and revise responses based on guiding principles, followed by AI-driven feedback to ensure transparency and safety in their outputs.

### 8.1.3 GEMINI
Google Gemini [25], announced in December 2023 as the successor to LaMDA and PaLM 2, is a family of multimodal large language models developed by Google DeepMind. Led by Sundar Pichai and Demis Hassabis, Gemini processes various data types like text, images, audio, video, and code. Launched with three versions—Ultra, Pro, and Nano—Gemini was designed to compete with models like GPT-4, with Ultra reportedly surpassing them on industry benchmarks. By early 2024, Gemini expanded into more applications, including Bard, Pixel smartphones, and Google Cloud's Vertex AI, with updates like Gemini 1.5 enhancing its capabilities and context window.

## 8.2 EVALUATION METRICS AND RESULTS

We compare four different metrics for the following LLMs: 1) Gemini, 2) ChatGPT, 3) Claude, and 4) RAG-powered LLM. We show that for our curated Q&A dataset, our model outperforms all other commercial LLMs in terms of those metrics. We briefly discuss each of the metrics used for our experimental validation purposes below:

**AlignScore:** AlignScore [26] is a holistic metric that evaluates factual inconsistencies like contradictions and hallucinations in text. By assessing the information alignment between a reference and candidate text, it provides a more comprehensive measure. Notably, AlignScore significantly outperforms ChatGPT and GPT-4 centric metrics across various benchmark datasets, demonstrating its superior effectiveness.

**ParaScore:** ParaScore [27] emerged from revisiting the automatic paraphrase evaluation process, ultimately challenging conventional wisdom by suggesting that a reference is not necessary for more effective paraphrase quality assessment. By explicitly modeling lexical divergence, ParaScore is a strong candidate for evaluating paraphrases. We use this metric based on the premise that, given a ground truth or reference answer for a specific question, the answer generated by any LLM-driven architecture should be a paraphrased version of the original answer.

**BERTScore:** BERTScore [28] is a metric that evaluates the similarity of each token in the candidate sentence to each token in the reference sentence. Rather than relying on exact matches, it calculates token similarity using the contextual embeddings of those tokens.

**ROUGE Score:** ROUGE [29] , short for Recall-Oriented Understudy for Gisting Evaluation, quantifies the overlap of units like n-grams, word sequences, and word pairs between a computer-generated summary and human-created ideal summaries. Originally developed to compare summaries generated by different models, this metric operates on the principle that candidate summaries should closely resemble reference summaries when language generation is accurate. Consequently, our RAG-powered LLM generated answers should achieve higher ROUGE scores if they closely match the ground truth. Table 6 presents the results for all four metrics for our RAG-powered GPT, compared with the other four baselines. As shown, our RAG-powered GPT clearly outperforms all the baseline models. Although 'Claude' has a higher recall value, our model leads overall in the F1-score.

**Table 6: Comparison of RAG-powered GPT with State-of-the-Art Commercial LLMs**

| Model | AlignScore | ParaScore | BERTScore | | | ROUGEL Score | | |
|---|---|---|---|---|---|---|---|---|
| | | | Precision | Recall | F1-Score | Precision | Recall | F1-Score |
| ChatGPT3.5 | 0.64651 | 0.64087 | 0.83419 | 0.86653 | 0.84965 | 0.19754 | 0.42210 | 0.25063 |
| ChatGPT4o | 0.71040 | 0.62080 | 0.85262 | 0.86084 | 0.85644 | 0.24638 | 0.41660 | 0.29307 |
| Gemini | 0.66220 | 0.57926 | 0.83840 | 0.84516 | 0.84147 | 0.21843 | 0.55644 | 0.29974 |
| Claude | 0.66579 | 0.66195 | 0.84620 | **0.89018** | 0.86737 | 0.25837 | 0.32660 | 0.25692 |
| RAG-powered GPT | **0.73201** | **0.70433** | **0.85312** | 0.88861 | **0.87033** | **0.32576** | **0.56292** | **0.37340** |

The advantages of the 'State-Wise Index' method are explained through experimental analysis, with the results presented in Table 7. The state-wise individual retriever-based method addresses key engineering aspects of our RAG pipeline development. We have utilized the llamaindex library as an integral tool in this process. When a query references one or more states (e.g., "What are the maximum penalties for failing to follow the data breach notification statutes in Ohio and Oklahoma?"), it may require retrieving and analyzing relevant segments from each state's legislative code. The state-wise retriever generates jurisdiction-specific responses, which are then compiled into a single, comprehensive answer. As shown in Table 7, the 'State-Wise Index' retriever—despite involving multiple response generations and subsequent aggregation—shows no significant difference in response time when compared to the 'WDI' method. While generating and merging multiple outputs for multi-document queries could theoretically increase latency, the observed impact on response time is negligible.

The two methods we developed, 'Whole Data Index' (WDI) and 'State-Wise Index' (SWI), are designed to address different use cases. As highlighted in Table 7, the SWI method proves more effective at identifying relevant chunks with higher precision when the query pertains to multiple states. By optimizing information retrieval and selecting the most pertinent data, SWI simplifies the generator's task. Since the relevance of the identified chunks directly influences the quality of the generated response, queries involving multiple states must be matched with the appropriate contexts. In contrast, WDI is sufficient for simpler queries concerning one or two states. Although the overhead of using SWI is marginally higher than WDI, users may opt to consistently use SWI, regardless of the query's complexity. While WDI offers faster performance for single-state queries due to its centralized structure, SWI achieves greater accuracy and efficiency for multi-state queries by utilizing localized indexes that enhance lookup relevance. Despite a slightly higher response overhead, SWI provides more accurate results than WDI in scenarios requiring access to multiple states. Depending on the query type, directing it to the appropriate index-based response generator ensures higher accuracy, even if it incurs more processing time compared to the simpler WDI approach.

**Table 7: Comparison of Response Time Between WDI and SWI**

| Query Type | Response Time (Seconds) | |
|---|---|---|
| | Method 'WDI' | Method 'SWI' |
| **Single Document Query:** "What is the definition of identification documents according to Alabama acts?" | 16.21 | **14.66** |
| **Multi-Document Query:** "Give me an extensive comparison between Colorado Transportation act and Connecticut transportation act." | **10.72** | 13.25 |
| **Complex Query:** "Give me a comparison between the Digital Crime Acts of Florida and its neighboring states." | 18.03 (States are not correctly identified) | 23.12 (States are correctly identified) |

## 9. CONCLUSIONS

These findings suggest that even state-of-the-art LLMs face challenges when responding to specialized queries, particularly those involving recently enacted legislation. This limitation is significant, as inaccurate outputs can misidentify legislative gaps and complicate the work of legal and policy stakeholders. Nonetheless, the results point to a promising direction—LLMs, when paired with recent legal developments, show clear potential in identifying gaps within the transportation cybersecurity landscape. The takeaway is straightforward: RAG-powered LLMs offer a strong foundation for detecting legislative deficiencies, with the added benefit of grounding responses in factual accuracy—an essential condition for meaningful legal analysis.

Addressing legislative gaps in connected and automated transportation systems is essential in a technological environment defined by rapid change. As public trust is central to the adoption of these systems, regulatory updates and new statutory frameworks are necessary to ensure their secure and responsible development. Key areas of focus include identifying underexamined issues and prioritizing cybersecurity and data privacy. The broader goal is to support adaptive legislation, reduce reliance on manual review, and enable smoother integration of emerging technologies. Our RAG-powered LLM framework supports this effort by enabling both concept-driven and state-specific analysis, helping to surface missing legislative elements across jurisdictions.

The constantly evolving nature of the legal landscape demands analysis grounded in the most current information available. Recent advances in generative AI have opened the door to producing informed responses within specialized domains by leveraging the capabilities of large language models. While LLMs carry known limitations, the integration of RAG introduces a factual foundation that makes the approach well-suited for critical applications such as legislative analysis. Our framework addresses the hallucination issues commonly observed in commercial LLMs when responding to queries about newly enacted laws. In this context, it offers a strong starting point for identifying legislative gaps in transportation cybersecurity.

## 10. FUTURE ADVANCEMENTS

Comprehensive methods can be developed to ensure factual accuracy in the responses generated for our dataset. These methods can scrutinize both questions and outputs for adherence to standards. From our observations, we have seen that the same query can yield different versions of answers given the same set of references or retrieved nodes, which are used for response generation via the RAG pipeline. Distillation of answers using the LLMs and extracting the best response from multiple sources can be an effective way to deal with such an issue. Afterwards, metrics like 'AlignScore' [26] will check for the enhancement of the factual basis and verify responses from the RAG pipeline. Methods such as ParaScore [27], BERTScore [28], and ROUGE score [29] can be also used to evaluate the similarity of the generated responses to the ground truths, further showcasing the effectiveness of the RAG-based framework in these challenging cases. We are currently in the process of launching the tool for public use, with a primary focus on collecting user feedback to improve its performance in real-world settings. As part of the framework's extension, we aim to enable both legal and technical users to upload relevant documents or information and receive accurate, context-aware answers to their queries. Additionally, we plan to automatically integrate newly published and amended state legislations into our vector database each year, ensuring the framework stays current. This scalability is essential to maintain the tool's long-term usability and relevance. To address these challenges, we are currently working on refining our validation process to be more domain-specific. In particular, we aim to ensure that factual representations adhere to legal standards, and to develop mechanisms for quantifying response correctness accordingly. We believe this direction will enhance both the reliability and interpretability of LLM-generated outputs in our target domain.

One limitation of this study is the absence of a dedicated module for comparing responses across different legislative bodies—an essential step in accurately identifying legislative gaps.

Addressing this will require significant curation of the Q&A dataset, along with targeted prompt engineering [30], to generate responses that meaningfully surface inconsistencies or omissions across jurisdictions. Questions related to the legal landscape may sometimes require guidance, prompting the querier to ask questions with specific keywords. We plan to address this in future work by incorporating features that guide users to ask questions with the right terms. The current version can answer free-style legal questions if they relate to the legal documents already stored in the vector database of our RAG system. Any new questions regarding enacted legislation can be easily addressed by appending the relevant information to the database, and the process is straightforward and seamless.

The scalability of our solution is a crucial factor in its utility. The computational and operational cost of maintaining a vector database in an RAG system largely depends on the need for reindexing. If the embedding model used to generate vector representations remains stable, the cost of adding new documents is generally linear and does not require full reindexing. However, periodic full reindexing may be necessary, particularly in domains where the source content is subject to regular updates—such as annual revisions to state legislation or corrections to existing laws. Although such reindexing incurs significant computational overhead, it is not required frequently and should be planned as part of scheduled maintenance cycles. In a nutshell, incremental document additions are computationally inexpensive as long as the embedding model remains stable. However, periodic full-scale reindexing may be warranted if previous reindexing efforts have demonstrated significant improvements in retrieval quality or model performance. Incremental updates involving thousands of documents can typically be completed within seconds [31], especially when using GPU acceleration. In contrast, full-scale reindexing may take minutes to hours [32], depending on the size of the dataset and the embedding model used, as it requires re-generating embeddings for all documents and rebuilding the entire vector index from scratch.

## 11. ACKNOWLEDGMENTS

This work is based upon the work supported by the National Center for Transportation Cybersecurity and Resiliency (TraCR) (a U.S. Department of Transportation National University Transportation Center) headquartered at Clemson University, Clemson, South Carolina, USA. Any opinions, findings, conclusions, and recommendations expressed in this material are those of the author(s) and do not necessarily reflect the views of TraCR, and the U.S. Government assumes no liability for the contents or use thereof. We have utilized 'ChatGPT4o' and 'ChatGPT3.5' to rephrase some of our own writing and to assist in summarizing the papers referenced in the 'Related Work' section.

## 12. AUTHOR CONTRIBUTIONS

The authors confirm their contributions to the paper as follows: Khandakar Ashrafi Akbar and Md Nahiyan Uddin contributed equally, with responsibilities including the design and assessment of the framework and the development of the RAG-powered LLM. All authors reviewed the results and approved the final version of the manuscript.

## 13. CONFLICT OF INTEREST

The authors declared no potential conflicts of interest with respect to the research, authorship, and/or publication of this article.

## 13. REFERENCES

1. Giannaros, A., Karras, A., Theodorakopoulos, L., Karras, C., Kranias, P., Schizas, N., Kalogeratos, G., & Tsolis, D. (2023). Autonomous Vehicles: Sophisticated Attacks, Safety Issues, Challenges, Open Topics, Blockchain, and Future Directions. Journal of Cybersecurity and Privacy, 3(3), 493-543. https://doi.org/10.3390/jcp3030025
2. Hockstad, Trayce & Rahman, Mizanur & Jones, Steven & Chowdhury, Mashrur. (2024). A regulatory gap analysis in transportation cybersecurity and data privacy. Transportation Journal. 64. 10.1002/tjo3.12036.
3. Balleisen, Edward J., and David A. Moss, eds. Government and Markets: Toward a New Theory of Regulation. Cambridge: Cambridge University Press, 2010.
4. Nick Goodson, Rongfei Lu, "Transforming Legal Aid with AI: Training LLMs to Ask Better Questions for Legal Intake", March 15, 2024, https://law.stanford.edu/2024/03/15/transforming-legal-aid-with-ai-training-llms-to-ask-better-questions-for-legal-intake/ (Date Accessed: December 1, 2024)
5. A3Logics, "How LLM's Are A Game Changer In Legal Research", Aug 5, 2024, https://www.a3logics.com/blog/llm-for-legal-research/ (Date Accessed: December 1, 2024)
6. GPT-4 Technical Report: https://cdn.openai.com/papers/gpt-4.pdf (Date Accessed: May 5, 2024)
7. Devlin, J., Chang, M.W., Lee, K., & Toutanova, K. (2019). BERT: Pre-training of Deep Bidirectional Transformers for Language Understanding. In Proceedings of the 2019 Conference of the North American Chapter of the Association for Computational Linguistics: Human Language Technologies, Volume 1 (Long and Short Papers) (pp. 4171–4186). Association for Computational Linguistics.
8. Hu, Y., Hosseini, M., Skorupa Parolin, E., Osorio, J., Khan, L., Brandt, P., & D′Orazio, V. (2022). ConfliBERT: A Pre-trained Language Model for Political Conflict and Violence. In Proceedings of the 2022 Conference of the North American Chapter of the Association for Computational Linguistics: Human Language Technologies (pp. 5469–5482). Association for Computational Linguistics.
9. Sun, H., Xu, G., Deng, J., Cheng, J., Zheng, C., Zhou, H., Peng, N., Zhu, X., & Huang, M. (2021). On the Safety of Conversational Models: Taxonomy, Dataset, and Benchmark. ArXiv. /abs/2110.08466
10. Zhang, Y., Li, Y., Cui, L., Cai, D., Liu, L., Fu, T., Huang, X., Zhao, E., Zhang, Y., Chen, Y., Wang, L., Luu, A. T., Bi, W., Shi, F., & Shi, S. (2023). Siren's Song in the AI Ocean: A Survey on Hallucination in Large Language Models. ArXiv. /abs/2309.01219
11. Huang, L., Yu, W., Ma, W., Zhong, W., Feng, Z., Wang, H., Chen, Q., Peng, W., Feng, X., Qin, B., & Liu, T. (2023). A Survey on Hallucination in Large Language Models: Principles, Taxonomy, Challenges, and Open Questions. ArXiv. /abs/2311.05232
12. Dziri, N., Milton, S., Yu, M., Zaiane, O., & Reddy, S. (2022). On the Origin of Hallucinations in Conversational Models: Is it the Datasets or the Models? ArXiv. /abs/2204.07931
13. Huo, S., Arabzadeh, N., & Clarke, C. L. (2023). Retrieving Supporting Evidence for Generative Question Answering. ArXiv. https://doi.org/10.1145/3624918.3625336

14. Liu, T., Zhang, Y., Brockett, C., Mao, Y., Sui, Z., Chen, W., & Dolan, B. (2021). A Token-level Reference-free Hallucination Detection Benchmark for Free-form Text Generation. ArXiv. /abs/2104.08704
15. Gao, T., Yen, H., Yu, J., & Chen, D. (2023). Enabling Large Language Models to Generate Text with Citations. ArXiv. /abs/2305.14627
16. "Hallucinating Law: Legal Mistakes with Large Language Models Are Pervasive." Stanford HAI, https://hai.stanford.edu/news/hallucinating-law-legal-mistakes-large-language-models-are-pervasive (Date Accessed: Aug 1, 2024)
17. Patrick Lewis, Ethan Perez, Aleksandra Piktus, Fabio Petroni, Vladimir Karpukhin, Naman Goyal, Heinrich Küttler, Mike Lewis, Wen-tau Yih, Tim Rocktäschel, Sebastian Riedel, & Douwe Kiela. (2021). Retrieval-Augmented Generation for Knowledge-Intensive NLP Tasks
18. Zhang, T., Patil, S. G., Jain, N., Shen, S., Zaharia, M., Stoica, I., & Gonzalez, J. E. (2024). RAFT: Adapting Language Model to Domain Specific RAG. ArXiv. /abs/2403.10131
19. "LlamaIndex." LlamaIndex Documentation, https://docs.llamaindex.ai/en/stable/ (Date Accessed: August 1, 2024)
20. Lewis, P., Perez, E., Piktus, A., Petroni, F., Karpukhin, V., Goyal, N., Küttler, H., Lewis, M., Yih, W., Rocktäschel, T., Riedel, S., & Kiela, D. (2021). Retrieval-Augmented Generation for Knowledge-Intensive NLP Tasks. arXiv preprint arXiv:2005.11401. Retrieved from https://arxiv.org/abs/2005.11401
21. https://openai.com/index/gpt-4o-and-more-tools-to-chatgpt-free/ (Date Accessed: May 5, 2024)
22. https://www.anthropic.com/news/claude-3-family (Date Accessed: May 5, 2024)
23. Wikipedia contributors. (2024, July 26). ChatGPT. In Wikipedia, The Free Encyclopedia. Retrieved 21:02, July 26, 2024, from https://en.wikipedia.org/w/index.php?title=ChatGPT-&oldid=1236692130
24. Wikipedia contributors. (2024, May 7). Claudia. In Wikipedia, The Free Encyclopedia. Retrieved 21:07, July 26, 2024, from https://en.wikipedia.org/w/index.php?title=Claudia-&oldid=1222632185
25. Wikipedia contributors. (2024, July 23). Gemini (language model). In Wikipedia, The Free Encyclopedia. Retrieved 21:11, July 26, 2024, from https://en.wikipedia.org/w/index.php?-title=Gemini_(language_model)&oldid=1236223829
26. Zha, Yuheng & Yang, Yichi & Li, Ruichen & Hu, Zhiting. (2023). AlignScore: Evaluating Factual Consistency with A Unified Alignment Function. 11328-11348. 10.18653/v1/2023.acl-long.634.
27. Shen, L., Liu, L., Jiang, H., & Shi, S. (2022). On the Evaluation Metrics for Paraphrase Generation. In Y. Goldberg, Z. Kozareva, & Y. Zhang (Eds.), Proceedings of the 2022 Conference on Empirical Methods in Natural Language Processing (pp. 3178–3190). Association for Computational Linguistics. https://doi.org/10.18653/v1/2022.emnlp-main.208
28. Zhang, T., Kishore, V., Wu, F., Weinberger, K. Q., & Artzi, Y. (2020). BERTScore: Evaluating Text Generation with BERT. 8th International Conference on Learning Representations, ICLR 2020, Addis Ababa, Ethiopia, April 26-30, 2020. https://openreview.net/forum?id=SkeHuCVFDr

29. Lin, C.-Y. (2004). ROUGE: A Package for Automatic Evaluation of Summaries. Text Summarization Branches Out, 74–81. https://aclanthology.org/W04-1013
30. "OpenAI Platform." Prompt Engineering, https://platform.openai.com/docs/guides/prompt-engineering (Date Accessed: Aug 1, 2024)
31. https://www.meilisearch.com/blog/how-meilisearch-updates-a-millions-vector-embeddings-database-in-under-a-minute?utm_source=chatgpt.com (Date Accessed: May 5, 2024)
32. https://www.weka.io/learn/guide/ai-ml/vector-dabase/ (Date Accessed: May 5, 2024)